\documentclass[11pt]{article}
\usepackage{naacl2021}
\usepackage{graphicx}
\usepackage{xcolor}
\usepackage{tipa}
\definecolor{color1}{rgb}{0.4, 0., 0.} 
\definecolor{color2}{rgb}{0.6, 0.4, 1} 

\usepackage{times}
\usepackage{latexsym}
\usepackage{xcolor}
\usepackage{linguex}
\usepackage{rotating}
\usepackage{float}

\usepackage{subfiles}
\newcommand{\onlyinsubfile}[1]{#1}
\newcommand{\notinsubfile}[1]{}
\usepackage{xr}
\usepackage{hyperref} 
\usepackage{mathtools}
\usepackage{booktabs}
\usepackage[export]{adjustbox}

\usepackage{multirow} % for the tables.
\usepackage{blindtext}
\usepackage{url}
\usepackage{amsmath}

\usepackage{fancyhdr}
\usepackage{fancyheadings}
\usepackage[bottom]{footmisc}

\title{Ab Antiquo: Neural Proto-language Reconstruction}

\author{Carlo Meloni$\footnotemark[1]$  \textsuperscript{1} \;\;\; Shauli Ravfogel$\thanks{~~Equal contribution}$ \textsuperscript{2,3} \;\;\; Yoav Goldberg\textsuperscript{2,3} \\
\textsuperscript{1}Department of Comparative Language Science, University of Zurich \\
\textsuperscript{2}Computer Science Department, Bar Ilan University \\
\textsuperscript{3}Allen Institute for Artificial Intelligence \\
 {\tt carlomeloni@mail.tau.ac.il, \{shauli.ravfogel, yoav.goldberg\}@gmail.com}
 }

\date{}

\begin{document}
\renewcommand{\onlyinsubfile}[1]{}

\maketitle
\begin{abstract}
Historical linguists have identified regularities in the process of historic sound change. The comparative method utilizes those regularities to reconstruct proto-words based on observed forms in daughter languages. Can this process be efficiently automated?  We address the task of proto-word reconstruction, in which the model is exposed to cognates in contemporary daughter languages, and has to predict the proto word in the ancestor language. We provide a novel dataset for this task, encompassing over 8,000 comparative entries, and show that neural sequence models outperform conventional methods applied to this task so far. Error analysis reveals a variability in the ability of neural model to capture different phonological changes, correlating with the complexity of the changes. Analysis of learned embeddings reveals the models learn phonologically meaningful generalizations, corresponding to well-attested phonological shifts documented by historical linguistics. 

\end{abstract}

\section{Introduction}

%intro to historical linguistics + general question
Historical linguists seek to identify and explain the various ways in which languages change through time. Research in historical linguistics has revealed that groups of languages (language families) can often be traced into a common, ancestral language, a ``proto-language''. Large-scale lexical comparison of words across different languages enables linguists to identify \emph{cognates}: words sharing a common proto-word. Comparing cognates makes it possible to identify rules of phonetic historic change, and by back-tracing those rules one can identify the form of the proto-word, which is often not documented. That methodology is called \emph{the comparative method} \cite{Anttila}, and is the main tool used to reconstruct the lexicon and phonology of extinct languages. 
%Consider, for example, the similarity between the words for ``wheel'' in the different Indo-european languages: \begin{IPA}hw\=eol\end{IPA} (Old English), \begin{IPA}k\'uklos\end{IPA} (Ancient Greek), \begin{IPA}cakr\'a\end{IPA} (Sanskrit) and \begin{IPA}kolo\end{IPA} (Old Church Slavonic). Based on those similarities, historical linguists were able to reconstruct the Proto-Indo-European word *\begin{IPA}k\textsuperscript{w}\'ek\textsuperscript{w}los\end{IPA} from which the daughter languages' words derived. 
Inferring the form of proto-words from existing cognates in daughter languages is possible since historical sound changes within a language family are not random. Rather, the phonological change is characterized by \emph{regularities} that are the result of constraints imposed by the human articulatory and cognitive faculties \cite{Trask}. For example, we can find such regular change---commonly called ``systematic correspondence''---by looking at the evolution of the first phoneme of Latin's word for ``sky'':\footnote{The words as transcribed with International Phonetic Alphabet (IPA) characters.}

\begin{figure}[H]
    \centering
    \includegraphics[scale=0.1]{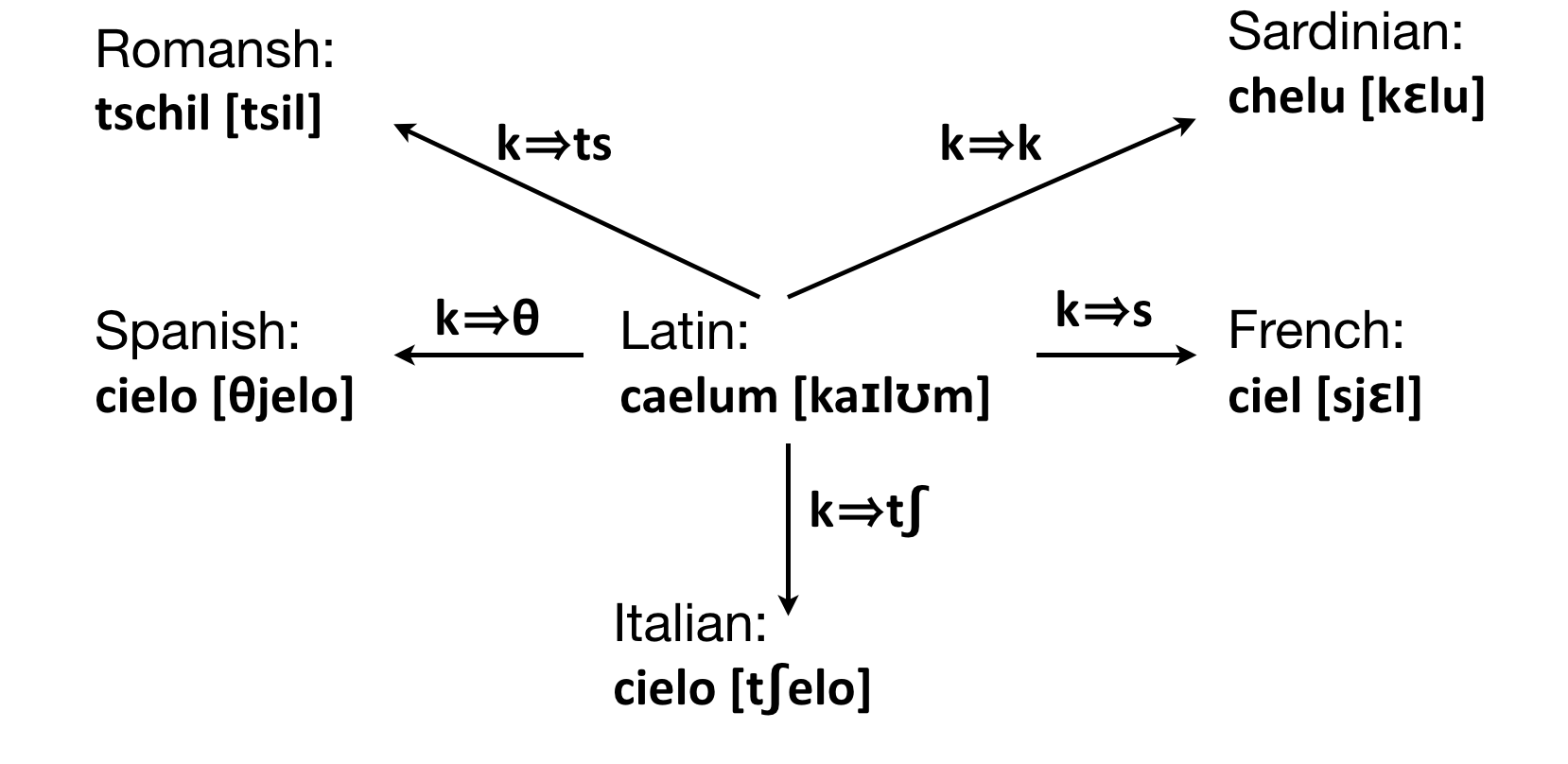}
        \caption{the evolution of Latin word for ``sky'' is several Romance languages.}    \label{plot:sky-romance}
\end{figure}

The Spanish word's first sound is \begin{IPA}[T]\end{IPA}, while the Italian word begins with \begin{IPA}[tS]\end{IPA}, the French word with \begin{IPA}[s]\end{IPA}, Romansh with \begin{IPA}[ts]\end{IPA} and Sardinian with \begin{IPA}[k]\end{IPA}. This pattern is systematic, and will be found throughout the languages. Working this way, historical linguists reconstruct words in the proto-language from existing cognates in the daughter languages, and determine how words in the proto-language may have sounded.

To what extent can a machine-learning model learn to reconstruct proto-words from examples in this way? And what generalizations of phonetic change will it learn? We focus on the task of proto-word reconstruction: the model is trained on sets of cognates and their known proto-word, and is then tasked with predicting the proto-word for an unseen set of cognates. Our study concentrate on the romance language family\footnote{All the languages that derived from Latin.} and the model is trained to reconstruct the Latin origin. We show that a recurrent neural-network model can learn to perform this task well (outperforming previous attempts).\footnote{We note that the role of the ML model is easier than that of the historical linguist, as it is trained on sets of words that it took the historical linguistics discipline a considerable effort to acquire.}

More interesting than the raw performance numbers are the learned generalizations and error patterns. 
The Romance languages are widely studied (\citet{ernst2003romanische, ledgeway2016oxford, holtus1989lrl} among others), and their phonological evolution from Latin is well mapped. The existence of this comprehensive knowledge allows exploring to what extent neural models internalize and capture the documented rules of language change, and where do they deviate from it. We provide an extensive error analysis, relating errors patterns to knowledge in historical linguistics. This is often not possible in common NLP tasks, such as parsing or semantic inference, in which the rules governing linguistic phenomena---or even the suitable framework to describe them---are still in dispute among linguists.

\paragraph{Contributions} Inspection of existing datasets of cognates in Romance languages has revealed inherent problems. We thus have collected a new comprehensive dataset for performing the reconstruction task (\S\ref{section-dataset}). Besides the dataset, our main contribution is the extensive analysis of what is being captured by the models, both on orthographic and phonetic versions of the dataset (\S\ref{section:error-analysis}). We find that the error patterns are not random, and they correlate with the relative opacity of the historic change. These patterns were divided in different categories, each one motivated by a sound phonological explanation. Moreover, in order to further evaluate the learning of rules of phonetic change, we evaluated models on a synthetic dataset (\S\ref{section:rule-validation}), showing that the model is able to correctly capture several phonological change rules. Finally, we analyze the learned inner representations of the model, and show it learns phonologically meaningful properties of phonemes (\S\ref{clustering}) and attributes different importance to different daughter languages (\S\ref{section-attention}).

\section{Related Work}

The related task of cognates detection has been extensively studied. In this task, a set of cognates should be extracted from word lists in different languages. Most effort in Machine learning approaches to this task has been focused on distance-based methods, which quantify the distance (according to some metric), or the similarity, between a given candidate of cognates. The similarity can be either static (e.g. Levenshtein distance) or learned. Once the metric is established, a classification can be performed either based on hard-decision (words below a certain threshold are considered cognates) or by learning a classifier over the distance measures and other features \cite{kondrak2001identifying, mann-distance-induction, inkpen2005automatic, ciobanu2014automatic, list2016using}; \citet{mulloni2006automatic} have evaluated an alternative approach, in which explicit rules of transformation are derived based on edit operations. See \citet{rama-cognates-eval} for a recent evaluation of the performance of several cognates detection algorithms. 

Several studies have gone beyond the stage of cognates extraction, and used resulted list of cognates to reconstruct the lexicon of proto-languages. Most studies in this direction borrowed techniques from computational phylogeny, drawing a parallel between the hypothesized branching of (latent) proto words into their (observed) current forms and the gradual change of genes during evolution. \citet{Bouchard2007} has applied such a model to the development of the Romance languages, based on a dataset composed of aligned-translations. \citet{bouchard2009improved, bouchard2013automated} used an extensive dataset of Austronesian languages and their reconstructed proto-languages, and built a parameterized graphical model which models the probability of a phonetic change between a word and its ancestral form; the probability is branch-dependent, allowing for the learning of different trends of change across lineages. While achieving impressive performance, even without necessitating a cognates lists as an input, their model is based on a given phylogeny tree that accurately represents the development of the languages in question.

\citet{DBLP:conf/lrec/WuY18} have automatically constructed cognate datasets for several languages, including Romance languages, and used a character-level NMT system to complete missing entries (not necessarily the proto-form). Several works studied the induction of multilingual dictionaries from partial data in related languages. \citet{DBLP:conf/lrec/WuNY20} reconstruct cognates in Austronesian languages (where the proto-language is not attested). \citet{DBLP:conf/coling/LewisWMY20} employ a mixture-of-experts approach for lexical translation induction, combining neural and probabilistic methods, and \citet{DBLP:journals/taslp/NishimuraSNN20} translate from a multi-source input that contains partial translations to different languages, concatenated.  Finally, \citet{ciobanu18} have applied a CRF model with alignment to a dataset of Romance cognates, created from automatic alignment of translations \cite{ciobanu-dataset}. The researchers also applied RNNs on the same dataset, but reported negative results.  

\section{Proto-word Reconstruction}
Our proto-word reconstruction is as follows: the training set is composed of pairs ($x_i$,$y_i$), where each $x_i=c_i^{\ell_1},...,c_i^{\ell_n}$ is a set of cognate words, each tagged with a language $\ell_j$, and $y_i$ is the proto-word (Latin word) of that set. 
We consider an \emph{orthographic task}, where the cognates and proto-words are spelled out as written.
As the orthography is often arbitrary and more conservative than spoken language, we consider also a \emph{phonetic task}, in which the cognates and proto-words are represented as their phonetic transcriptions into IPA.

An example of a training instance $(x,y)$ for the orthographic task is:

$x=$lapte$^{\textsc{Rm}}$,   
lait$^\textsc{Fr}$,
latte$^\textsc{It}$,
leche$^\textsc{Sp}$,
leite$^\textsc{Pt}$

$y=$lactem

\noindent and for the phonetic task is:

$x=$\begin{IPA}lapte\end{IPA}$^{\textsc{Rm}}$,
\begin{IPA}lE\end{IPA}$^{\textsc{Fr}}$,
\begin{IPA}latte\end{IPA}$^\textsc{It}$,
\begin{IPA}letSe\end{IPA}$^\textsc{Sp}$,
\begin{IPA}l5jt\textbari\end{IPA}$^\textsc{Pt}$

$y=$\begin{IPA}laktEm\end{IPA}

A cognate in one of the languages may be missing, in which case we represent it by a dash. Here, we are missing the Italian and Romanian cognates:
    
$x=$--$^{\textsc{Rm}}$,
\begin{IPA}tKavaj\end{IPA}$^{\textsc{Fr}}$,
--$^\textsc{It}$,
\begin{IPA}tRabaxo\end{IPA}$^\textsc{Sp}$,
\begin{IPA}tR5BaLu\end{IPA}$^\textsc{Pt}$

$y=$\begin{IPA}trIpalEm\end{IPA}

At test time, we are given a set of cognates and are asked to predict their proto-word.

\section{Comprehensive Romance Dataset}
\label{section-dataset}

The different experiments described in the paper were performed on a large dataset of our creation, which contained cognates and their proto-words in both orthographic and phonetic (IPA) forms. The dataset's departure point is \citet{ciobanu-dataset}, which consists of 3,218 complete cognate sets in six different languages: French, Italian, Spanish, Portuguese, Romanian and Latin \footnote{We thank Ciobanu and Dinu for sharing their data with us.}. We augmented the dataset's items with a freely available resource, Wiktionary, whose data were manually checked against \citet{etymologicalDict} to ensure their etymological relatedness with the Latin source. The entries were transcribed into IPA using the transcription module of the eSpeak library\footnote{https://github.com/espeak-ng/espeak-ng}, which offers transcriptions for all languages in our dataset, including Latin. The final dataset contains 8,799 cognate sets (not all of them complete), which were randomly splitted into train, evaluation and test sets: 7,038 cognate sets (80\%) were used for training, 703 (8\%) for evaluation and 1,055 (12\%) for testing. Overall, the dataset contains 41,563 distinct words for a total of 83,126 words counting both the orthographic and the phonetic datasets. Vowel lengths were found to be difficult to recover (see Table \ref{tbl:general-results}), hence we created the following variations of the dataset: with and without vowel length (for both the orthographic and phonetic datasets), and without a contrast (for the phonetic dataset); see section \S\ref{section:error-analysis} for further discussion.

A detailed description of the dataset collection process is available at the appendix \S \ref{appendix:dataset-creation}. We make our additions to the dataset of \citet{ciobanu-dataset} publicly available \footnote{\url{https://github.com/shauli-ravfogel/Latin-Reconstruction-NAACL}. The entries that appeared in the original dataset are not publicly available.}.

\begin{table*}[t]

  \resizebox{0.95\textwidth}{!}{\begin{minipage}{\textwidth}

\begin{tabular}{lccccccc}
\hline
\multicolumn{1}{c}{}          & \multicolumn{5}{c}{\textbf{Edit Distance}}                         & \multicolumn{1}{l}{}                 & \multicolumn{1}{l}{}                             \\
\multicolumn{1}{c}{}          & \textbf{0} & \textbf{$\leq$1} & \textbf{$\leq$2} & \textbf{$\leq$3} & \textbf{$\leq$4} & \multicolumn{1}{l}{\textbf{Average}} & \multicolumn{1}{l}{\textbf{Avg, norm}} \\ \hline
Ortographic & 64.1\%       & 84.0\%        & 92.7\%        & 96.8\%        & 98.5\%        & 0.65                                & 0.064                                          \\
IPA         & 50.0\%       & 73.9\%        & 85.9\%        & 93.3\%        & 97.0\%        & 1.022                                & 0.100                                           \\
\hline
Ortographic, added vowel lengths    & 42.5\%       & 72.0\%        & 86.3\%        & 92.5\%        & 96.7\%        & 1.13                                 & 0.102                                           \\
IPA, added vowel lengths            & 47.9\%       & 62.0\%        & 80.2\%        & 87.1\%        & 93.8\%        & 1.331                                & 0.119    \\

\hline
IPA, no contrast            & 65.1\%       & 80.0\%        & 87.5\%        & 93.6\%        & 96.7\%        & 0.797                                & 0.077

\end{tabular}
\caption{\label{tbl:general-results} Distribution of edit distances between the reconstructed and original Latin form, on the  orthographic and transcribed datsaets. Edit distance of 0 corresponds to perfect reconstruction. ``Average" refers to average edit distance, and ``Avg, norm" to normalized average edit distance.}

      \end{minipage}}

\end{table*}

\section{Experimental Setup}

\subsection{NMT-based Neural Model}

Our proto-word reconstruction setup follows an encoder-decoder with attention architecture, similar to contemporary neural machine translation (NMT) systems \cite{encoder-decoder-attention, bengio-NMT}.

We use a standard character-based encoder-decoder architecture with attention \cite{encoder-decoder-attention}. Both encoder and decoder are GRU networks with 150 cells. The encoder reads the forms of the words in the daughter languages, and output a contextualized representation of each character. At each decoding step, the decoder attends to the encoder's representations via a dot-product attention. The output of the attention is then fed into a MLP with 200 hidden units, which outputs the next Latin character to generate.

%Contrary to most previous works on the subject, which have devised a graphical model that represent the gradual transformation of the Latin word to its descendants in the different Romance languages, we frame the problem as learning a \emph{mapping} between the input, which constitutes of a list of cognates in the 5 Romance languages we use, and the output -- which is the Latin proto-word. This framing allows us to use a standard neural architecture, similar to the ones commonly used for tasks such as machine translation.

\paragraph{Input representation}
Each character (a letter in the orthographic case, and a phoneme in the phonetic case) is represented by an embedding vector of size 100. While all Romance languages are orthographically similar, the same letters represent different sounds, and thus convey different kinds of information for the task of Latin reconstruction. A possible approach would encode each language's characters using a unique embedding table. We instead share the character embedding table across all languages (including Latin), but concatenate to each character vector also a language-embedding vector.
The final representation of a character $c$ in language $\ell$ is then $W E[c] + U E[\ell]$ where $E$ is a shared embedding matrix, $c$ is a character id, $\ell$ is a language id, and $W$ and $U$ are a linear projection layers.

%by embedding vectors. While the orthography is similar among all Romance languages, the same letters represent different sounds -- and thus convey different kinds of information for the task of Latin reconstruction -- in different languages. As such, we also use language embedding vectors (one for each daughter language, and one for Latin). We concatenate the language-agnostic representation of the phoneme or the letter with the embedding of the language in which the character appeared, and pass it through a linear layer to get the final, language-dependent representation of a character. 

%\section{Reconstruction Task}

%For all experiments described here, we trained different models on the orthographic and phonetic datasets. In both settings, the model is exposed to the cognates in French, Italian, Romanian, Spanish and Portuguese, and has to predict the form of the Latin proto-word. Consider, for example, the Latin word ``laborare'' (``to work''). In the orthographic setting, the model is exposed to its descendants in the 5 Romance languages:

%\begin{exe}
%		\ex 
%        \gll Rm: - Fr: labourer It: lavorare Sp: laborar Pt: laborar 
%        \label{ex:cognates_list}
%\end{exe}
%
%And is expected to output the Latin ``laborare'. Note that in this example, the Romanian cognate was not included in the dataset, and is thus represented by ``-'.

\subsection{Evaluation Metric} Our main quantitative metric for evaluation is the edit distance between the reconstructed word and the gold Latin word. We use the standard edit distance with equal weight of 1 for deletion, insertion and substitution. We report test set average edit distance and average normalized edit distance (divided by word length), as well as the percentage of instances with less than $k$ edit operations between the reconstruction and the gold, for $k=0$ to $4$. 

%In addition to this quantitative measure, we perform a linguistic error analysis that focuses on the relative opacity of the various phonological changes that have occurred between Latin and its descendent languages (sections \ref{orto-analysis} and \ref{ipa-analysis}). 
%In addition, we evaluate the extent to which the model had learned the known rules of phonetic transformations of syllables between Latin and its daughter languages
%(\S\ref{section:rule-validation}).  

\section{Results and Analysis}
\label{section:error-analysis}
Table \ref{tbl:general-results} summarizes our main quantitative results. ``Orthographic, added vowel lengths'' and ``IPA, added vowel lengths'' refer to variations of the datasets that include  explicit marking of vowel length in Latin words, marked by $\textless$\begin{IPA}:\end{IPA}$\textgreater$ after long vowels.
The model’s performance on the orthographic dataset demonstrates a substantial improvement over previously reported results. Our method has achieved average edit distance of 0.65, average normalized edit distance of 0.064, and 64.1\% complete reconstruction rate (edit distance of 0). These numbers compare favorably with the edit distance of 1.07, normalized edit distance of 0.13 and 50\% complete reconstruction reported by \citet{ciobanu18}. We note, however, that as our method is different both in the training corpus and in the type of model we employ, it is not clear whether this improvement should be attributed to the quality of the data, to the model, or to both of them.\footnote{When we train a smaller version of our model (75-dimensional GRU) on the original dataset of \citet{ciobanu-dataset} we achieve average edit distance of 0.881, average normalized edit distance of 0.103, and complete reconstruction rate of 59.1\%. Training a similar model on their dataset after cleaning resulted in average edit distance of 0.612, average normalized edit distance of 0.062 and complete reconstruction rate of 68.8\%.} 

The performances on the phonetic dataset were lower than those derived from the orthographic one: in the phonetic dataset the average edit distance was of 1.022, and the average normalized edit distance of 0.1, with 50.0\% complete reconstruction rate. 

This disparity can be explained at least partially by a peculiarity of the phonetic dataset: it implicitly encodes vowel length, which was neutralized in the orthographic dataset. The reason for this difference is that length contrast in Latin co-occurred with quality differences: short vowels tended to be more open than their long counterparts, a contrast also called ``tense-lax'' \cite{LtOrtho}. This contrast is not present in Latin orthography, but it appears in its phonetic transcription. This results in a noticeable gap between the results of the orthographic dataset with vowel lengths and without vowel lenghts (0.064 average normalized edit distance vs. 0.119), while the differences between the phonetic IPA dataset with vowel lenghts and without vowels lengths are much smaller. When the contrast ``tense-lax'' is manually neutralized\footnote{We achieved that by respectively changing the characters $\textless$\begin{IPA}U\end{IPA}$\textgreater$, $\textless$\begin{IPA}O\end{IPA}$\textgreater$,
$\textless$\begin{IPA}I\end{IPA}$\textgreater$,
$\textless$\begin{IPA}E\end{IPA}$\textgreater$ to $\textless$\begin{IPA}u\end{IPA}$\textgreater$, $\textless$\begin{IPA}o\end{IPA}$\textgreater$, $\textless$\begin{IPA}i\end{IPA}$\textgreater$, $\textless$\begin{IPA}e\end{IPA}$\textgreater$ in the Latin words}, the performances achieved are similar to the ones on the orthographic dataset (as it is possible to see from the performances on ``IPA, no contrast'', whose Latin entries do not contain a ``tense-lax'' contrast).

%The main quantitative results on both the orthographic and the phonetic datasets are summarized in table \ref{tbl:general-results}.

\subsection{Error Patterns}
\label{section:error-patterns}

%\begin{table*}[t]

% \resizebox{0.7\textwidth}{!}{\begin{minipage}{\textwidth}

\begin{table}[t]

\begin{tabular}{lcc}
Error type & Orthographic & Phonetic \\
\hline
\textbf{High-mid} & 18\% & 8\% \\
\textbf{Deletion} & 14\% & 6\% \\
\textbf{Consonant} & 13\% & 15\% \\
\textbf{Cluster} & 12\% & 3\% \\
\textbf{Morphology} & 11\% & 10\% \\
\textbf{Vowel} & 7\%  & 8\% \\
\textbf{Length} & --- & 26\% \\
\textbf{Orthography} & 5\% & --- \\
\textbf{Other} & 20\% & 24\% \\
\end{tabular}

\caption{Error type distribution based on 650 orthographic and 650 phonological errors.}

\label{error-distribution}
\end{table}

%\input{error-distribution.tex}
% \caption{Main groups of errors (\%) on the phonetic and orthographic datasets}
%     \label{error-distribution}
%      \end{minipage}}
%\end{table*}

\begin{figure}
    \includegraphics[width=0.4\textwidth]{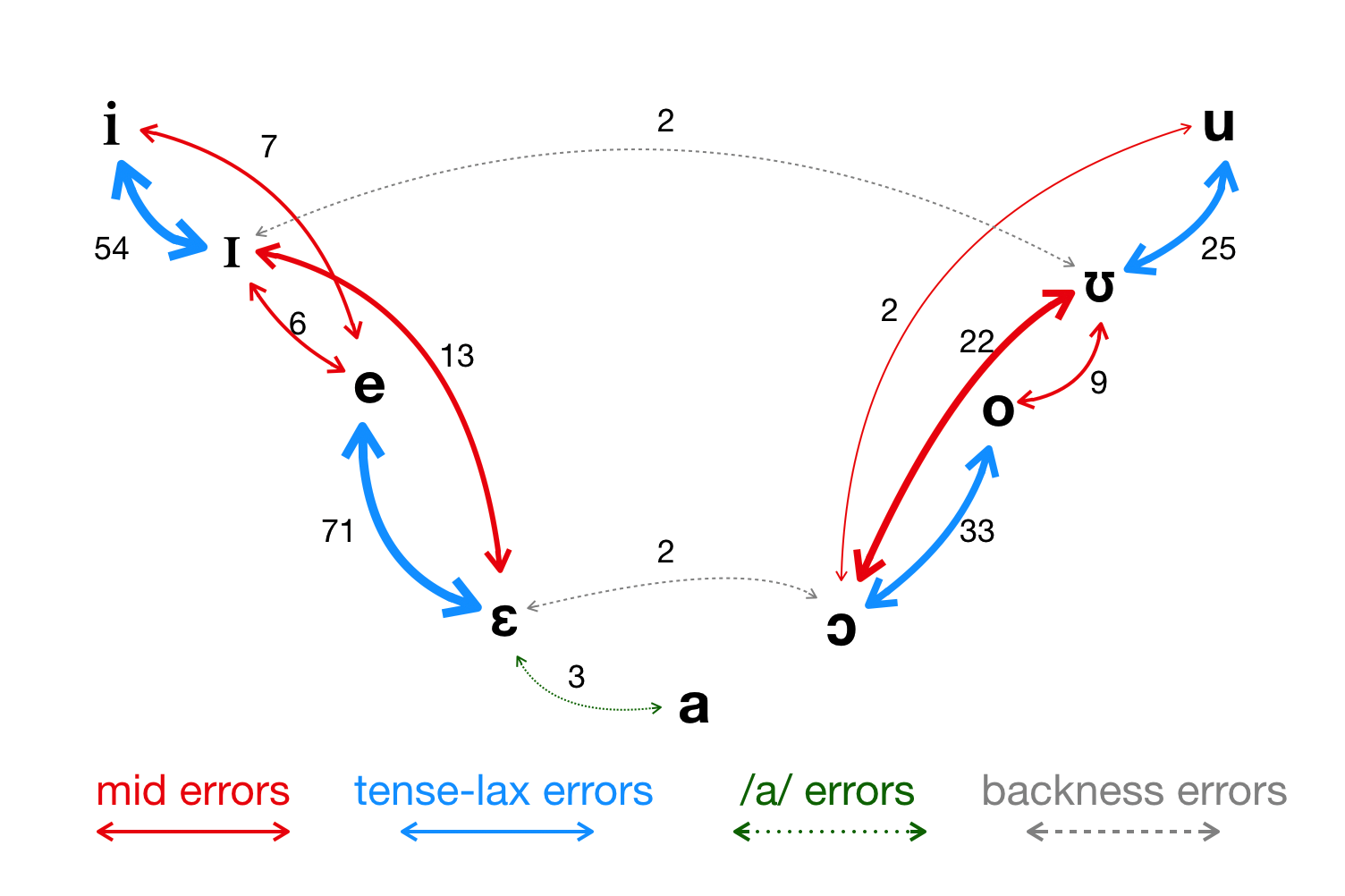}
    \vspace{-1.5em}
        \caption{Phonological mistakes resulting from alternations between \emph{vowels}, on the phonetic dataset. The numbers signify the number of errors, excluding singleton errors.}
    \label{plot:phon-errors}
\end{figure}

The following subsections focus on the model performances on the orthographic and phonetic datasets without explicit vowel length marking. A thorough analysis of both datasets reveals that the model's errors are not arbitrary, but rather tend to correspond to one of a few well-defined linguistic phenomena characterizing the evolution of Latin to its daughter languages. From an analysis of about 1300 errors, equally divided between the orthographic and the phonetic datasets, we find that 80\% of the errors of the model on the orthographic dataset, and 75\% on the phonetic one can be grouped into one of the following groups: high-mid vowel alternations, segment deletion, segment changes, cluster changes, morphological changes and other vowel changes.
Additionally, one error category is unique to the phonetic dataset, tense-lax errors, and one is unique to the orthographic dataset, orthography errors.
Table \ref{error-distribution} summarizes the results, and Figure \ref{plot:phon-errors} visualizes the vowels error patterns on the phonetic dataset.
 
We briefly discuss each of these groups. \footnote{The orthographic characters will be displayed between two angle brackets, while phonetic characters between two square brackets.}

\noindent\textbf{High-mid alternation.} The largest number of errors on the orthographic dataset, 18\%, can be attributed to confusion between high and mid-high vowels (correspondingly $\textless$i$\textgreater$, $\textless$u$\textgreater$ and $\textless$e$\textgreater$, $\textless$o$\textgreater$), as shown by the reconstruction $\textless$pescarium$\textgreater$ instead of the Latin $\textless$piscarium$\textgreater$ (alternation between $\textless$e$\textgreater$ and $\textless$i$\textgreater$). That error is much rarer in the phonetic dataset, accounting only for 8\% of all the errors. The reason of this error can be attributed to the origin of the mid-vowels in the daughter languages: while Latin long vowels \begin{IPA}[i:]\end{IPA}, \begin{IPA}[e:]\end{IPA}, \begin{IPA}[o:]\end{IPA} and \begin{IPA}[u:]\end{IPA}, always evolved into \begin{IPA}[i]\end{IPA}, \begin{IPA}[e]\end{IPA}, \begin{IPA}[o]\end{IPA} and \begin{IPA}[u]\end{IPA} in the daughter languages (with minor changes related to syllable structure), Latin short vowels---\begin{IPA}[I]\end{IPA}, \begin{IPA}[E]\end{IPA}, \begin{IPA}[O]\end{IPA} and \begin{IPA}[U]\end{IPA}---are not deterministically mapped into corresponding vowels in the daughter languages: Latin \begin{IPA}[I]\end{IPA} and \begin{IPA}[E]\end{IPA} both usually became Romance [e] (with alternations related to syllable structure, as diphthongization to [je]), while Latin \begin{IPA}[O]\end{IPA} and \begin{IPA}[U]\end{IPA} have different reflexes in the daughter languages as [u], [o], \begin{IPA}[O]\end{IPA}, \begin{IPA}[\o]\end{IPA} or as diphthongs. Because of this complex evolution, which merges different Latin phonemes into the same one in the daughter languages, the model is unable of unequivocally predicting the Latin vowel. Nonetheless, it seems that the tense-lax contrast present in the phonetic dataset eases the task of distinguishing the different phonemes, and enables the network to reconstruct their origin more often.\\
\noindent \textbf{Segment deletion.} examples of these errors are the reconstruction of $\textless$aspargum$\textgreater$ instead of Latin $\textless$asparagum$\textgreater$, and the reconstruction of \begin{IPA}[abIlItatEm]\end{IPA} instead of Latin \begin{IPA}[habIlItatEm]\end{IPA}. During the evolution from Latin to Romance languages, unstressed syllables tended to be dropped. This phenomenon was not systematic, and occurred in different ways among and within the languages. Such process could affect either whole syllables (consonant + vowel) or only the vowel, creating new consonant clusters. Because of the erratic nature of this process, it seems that the network struggles with the exact reconstruction of segments eliminated in the daughter languages. A special kind of deletion is that of the consonant \begin{IPA}[h]\end{IPA}. This consonant did not survive in any Romance languages (although it may be represented orthographically), and hence many times the network does not reconstruct it. \\
\noindent \textbf{Segment changes.} this category encompasses errors in the reconstruction of consonants--- such as voicing changes (reconstructing  $\textless$faculdadem$\textgreater$ vs. Latin $\textless$facultatem$\textgreater$), assimilation (\begin{IPA}[wessarE]\end{IPA} vs. \begin{IPA}[weksarE]\end{IPA}) and gemination (\begin{IPA}[agrEgatIonEm]\end{IPA} vs. \begin{IPA}[aggrEgatIonEm]\end{IPA}). All these errors reflect processes that took place in all of the daughter languages, that obscures the original form of the proto-word. \\
\noindent \textbf{Cluster changes.} These are changes that occur with two contiguous consonants. Consider, for example, the reconstruction of \begin{IPA}[rEatIonEm]\end{IPA} instead of Latin \begin{IPA}[rEaktIonEm]\end{IPA}, and of $\textless$sennorem$\textgreater$ instead of Latin $\textless$seniorem$\textgreater$. The former is an instance of cluster simplification, while the latter is an instance of cluster palatalization. In many of the daughter languages clusters of two different sounds underwent simplification, either by the dropping of one of the sound or the assimilation of one of them. Palatalization is the process by which certain sounds tend to be pronounced more closely to the palate, usually because of an adjacent front vowel. This change occurred in all Romance languages, even though its orthographic representation may vary among them.\\
\noindent \textbf{Morphological changes.} Latin had a very developed morphology, with several classes of special conjugations and irregular forms. The network struggles to reconstruct correctly irregular forms, as these forms were mostly lost in the daughter languages. An instance of such irregular verbs is $\textless$praeferre$\textgreater$, reconstructed as $\textless$praeferire$\textgreater$ by the network. Moreover, other special morphological classes, such as Latin neuters, tend to be reconstructed as more usual forms. Another interesting class of errors is change of morphological category: some nouns have suffixes reminiscent of those of verbs, and hence are wrongly reconstructed as such.
A separated case is that of Greek words: Latin contained several Greek loanwords that conserved their original morphology, different from the Latin one. Since these peculiarities were, for most part, not retained in the daughter languages, the network reconstructs them with normal Latin suffixes. For example, the greek \begin{IPA}[syntaksIs]\end{IPA} was reconstructed as \begin{IPA}[syntaksEm]\end{IPA}, with the normal Latin suffix.\\
\noindent \textbf{Other vowel changes.} Latin contained several diphthongs, among them the diphthongs \begin{IPA}[aI]\end{IPA} and \begin{IPA}[OI]\end{IPA}. These sounds did not survive in any of the daughter languages (although in some rare cases they may be represented in the orthography), and both changed into \begin{IPA}[e]\end{IPA} in the different Romance languages. This lef to reconstruction errors such as reconstructing $\textless$egrum$\textgreater$ instead of Latin $\textless$aegrum$\textgreater$. Some changes also occurred with the vowel \begin{IPA}[a]\end{IPA}, which was reconstructed as a different vowel.\\
\noindent \textbf{Greek orthography.} some Latin words from Greek origin retained some orthographic conventions alien to Latin, such as the use of $\textless$y$\textgreater$, $\textless$ph$\textgreater$, $\textless$th$\textgreater$, $\textless$rh$\textgreater$ etc. These conventions were only partially retained in the daughter languages, which creates some inconsistencies in their reconstruction by the network.\\
\noindent \textbf{Tense-Lax alternation.} this is the largest category found in the network’s errors on the phonetic dataset -- up to 26\% of all errors. As said previously, the tense-lax contrast reflects vowel length in Latin, which is not entirely predictable based on the daughter languages. The network tends to confuse between the lax and the tense vowels. 

Figure \ref{plot:phon-errors} shows clearly that the network's errors are internally consistent and not random: all the vowel errors fit neatly in one of the aforementioned categories, while other possible errors do not occur.

\paragraph{Orthographic vs. Phonetic} Importantly, the phonetic and orthographic tasks differ in their error distributions: while the performance of the network on the orthographic task displays many syllable changes -- changes that alter the structure of the syllable (mostly changes in consonant clusters and deletion of segments) -- on the phonetic tasks the model tends to retain syllable structure, but perform more segment-related errors (i.e., changing a specific vowel or consonant for another one). The IPA performance contained more idiosyncratic errors that could not be categorized in one of the main categories. Such errors tended to occur when the network had only one or two cognates from the daughter languages. Even though the orthographic performance also exhibited poorer reconstructions in these cases, it seems that the IPA performance was even more affected by the singular words, leading to more erratic reconstructions.

\subsection{Learnt generalizations}

% \begin{figure*}[t]
%     %\vspace{-4em}
%     \includegraphics[0.2\textwidth]{clusts-ward-f.png} 
%     \vspace{-4.5em}
%     \caption{Hierarchical clustering of learned phoneme representations for all languages.}
%     \label{clust-all}
%       %\end{minipage}
% \end{figure*}

This section will focus on the phonetic dataset. A closer inspection of the errors made by the model, and of those that do not occur in the data, can shed light on the processes of phonological change learnt by the model. We will first focus on the vowels. The Latin vowel \begin{IPA}[a]\end{IPA} is quite resilient to changes, and most of the daughter languages retain it without change (only in French and Romanian some phonological changes occur, in certain phonological environments). Indeed, the network has almost no mistakes in recovering it, apart from some isolated cases that derives from insufficient cognates in the daughter languages. The network also makes virtually no errors regarding the reconstruction of vowel backness -- here also the only few cases are caused by the paucity of cognates and by assimilation processes in the daughter languages that make the Latin source opaque (metaphony processes). All in all, the network learns correctly the phonological changes that occurred in Latin vowels, and the main errors are a result of changes that cannot be fully reverted from the daughter languages. 

The model learnt well the mapping of consonants between Latin and its daughter languages, and vowel reconstruction errors are considerably more prevalent. Focusing on one type of errors, palatalization, shows that the network failed to reconstruct the original consonant in opaque contexts, that is, when phonological cues crucial for the right reconstruction were lacking. Specifically, the network confused between the consonants \begin{IPA}[t]\end{IPA} and \begin{IPA}[k]\end{IPA} in the Latin reconstruction, since they palatalize to the same segments in Spanish and French. Without the other daughter languages, it is impossible to reconstruct correctly the original sound in Latin.

Finally, the network correctly generalized the occurrence of nasals in Latin clusters. Latin nasal tended to assimilate to the place of articulation of the adjacent consonant, deriving clusters such \begin{IPA}[Nk]\end{IPA}, \begin{IPA}[mp]\end{IPA} and \begin{IPA}[nt]\end{IPA}. When the network deleted a consonant in a cluster containing a nasal, or changed a consonant adjacent to a nasal, the nasal consonant always changed to match the place of articulation of the following consonant. Hence, by deleting \begin{IPA}[k]\end{IPA} in the cluster \begin{IPA}[Nkt]\end{IPA}, the network reconstructs \begin{IPA}[nt]\end{IPA}. Similarly, by changing \begin{IPA}[p]\end{IPA} to \begin{IPA}[t]\end{IPA} in the cluster \begin{IPA}[mp]\end{IPA}, the nasal consonant accordingly: \begin{IPA}[nt]\end{IPA}.

\subsection{Evaluating Rules of Phonetic Change} 
\label{section:rule-validation}

\begin{table*}[t]

\centering
\begin{scalebox}{0.8}{
\begin{tabular}{lllllllll}

\textbf{Latin phoneme} & \textbf{Romanian} & \textbf{French} & \textbf{Italian} & \textbf{Spanish} & \textbf{Portuguese} & \textbf{Latin}  & \textbf{Latin - reconstruction} & \textbf{Correct}\\ \hline
\begin{IPA}/e/\end{IPA} blocked syllable & \begin{IPA}p\textbf{e}p\end{IPA} & \begin{IPA}p\textbf{e}p\end{IPA} & \begin{IPA}p\textbf{e}p\end{IPA} & \begin{IPA}p\textbf{e}p\end{IPA} & \begin{IPA}p\textbf{e}p\end{IPA} & \begin{IPA}p\textbf{e}p\end{IPA} & \begin{IPA}p\textbf{I}p\end{IPA} & no \\
\begin{IPA}/o/\end{IPA} blocked syllable & \begin{IPA}p\textbf{o}p\end{IPA} & \begin{IPA}p\textbf{u}p\end{IPA} & \begin{IPA}p\textbf{o}p\end{IPA} & \begin{IPA}p\textbf{o}p\end{IPA} & \begin{IPA}p\textbf{o}p\end{IPA} & \begin{IPA}p\textbf{o}p\end{IPA} & \begin{IPA}p\textbf{U}p\end{IPA} & no \\
\begin{IPA}/E/\end{IPA} blocked syllable & \begin{IPA}p\textbf{je}p\end{IPA} & \begin{IPA}p\textbf{E}p\end{IPA} & \begin{IPA}p\textbf{E}p\end{IPA} & \begin{IPA}p\textbf{je}p\end{IPA} & \begin{IPA}p\textbf{E}p\end{IPA} & \begin{IPA}p\textbf{E}p\end{IPA} & \begin{IPA}p\textbf{e}p\end{IPA} & no \\
\begin{IPA}/kt/\end{IPA} medially, before nasals & - & \begin{IPA}an\textbf{t}a\end{IPA} & \begin{IPA}an\textbf{t}a\end{IPA} & \begin{IPA}an\textbf{t}a\end{IPA} & \begin{IPA}an\textbf{t}a\end{IPA} & \begin{IPA}an\textbf{kt}a\end{IPA} & \begin{IPA}an\textbf{t}am\end{IPA} & no \\
\begin{IPA}/aI/\end{IPA} & \begin{IPA}p\textbf{e}\end{IPA} & \begin{IPA}p\textbf{e}\end{IPA} & \begin{IPA}p\textbf{e}\end{IPA} & \begin{IPA}p\textbf{e}\end{IPA} & \begin{IPA}p\textbf{e}\end{IPA} & \begin{IPA}p\textbf{aI}\end{IPA} & \begin{IPA}p\textbf{E}m\end{IPA} & no \\
\begin{IPA}/OI/\end{IPA} & \begin{IPA}p\textbf{e}\end{IPA} & \begin{IPA}p\textbf{e}\end{IPA} & \begin{IPA}p\textbf{e}\end{IPA} & \begin{IPA}p\textbf{e}\end{IPA} & \begin{IPA}p\textbf{e}\end{IPA} & \begin{IPA}p\textbf{OI}\end{IPA} & \begin{IPA}p\textbf{E}m\end{IPA} & no \\
\begin{IPA}/b/\end{IPA} intervocalic & \begin{IPA}aa\end{IPA} & \begin{IPA}a\textbf{v}a\end{IPA} & \begin{IPA}a\textbf{v}a\end{IPA} & \begin{IPA}a\textbf{B}a\end{IPA} & \begin{IPA}a\textbf{v}a\end{IPA} & \begin{IPA}a\textbf{b}a\end{IPA} & \begin{IPA}a\textbf{w}am\end{IPA} & no \\
\begin{IPA}/e/\end{IPA} free syllable & \begin{IPA}p\textbf{e}\end{IPA} & \begin{IPA}p\textbf{wa}\end{IPA} & \begin{IPA}p\textbf{e}\end{IPA} & \begin{IPA}p\textbf{e}\end{IPA} & \begin{IPA}p\textbf{e}\end{IPA} & \begin{IPA}p\textbf{e}\end{IPA} & \begin{IPA}p\textbf{E}m\end{IPA} & no \\
\begin{IPA}/o/\end{IPA} free syllable & \begin{IPA}p\textbf{o}\end{IPA} & \begin{IPA}p\textbf{\o}\end{IPA} & \begin{IPA}p\textbf{o}\end{IPA} & \begin{IPA}p\textbf{o}\end{IPA} & \begin{IPA}p\textbf{o}\end{IPA} & \begin{IPA}p\textbf{o}\end{IPA} & \begin{IPA}p\textbf{U}m\end{IPA} & no \\
\begin{IPA}/I/\end{IPA} free syllable & \begin{IPA}p\textbf{e}\end{IPA} & \begin{IPA}p\textbf{wa}\end{IPA} & \begin{IPA}p\textbf{e}\end{IPA} & \begin{IPA}p\textbf{e}\end{IPA} & \begin{IPA}p\textbf{e}\end{IPA} & \begin{IPA}p\textbf{I}\end{IPA} & \begin{IPA}p\textbf{E}m\end{IPA} & no \\
\begin{IPA}/n/\end{IPA} before front vowels & \begin{IPA}\textbf{j}i\end{IPA} & \begin{IPA}\textbf{\textltailn}i\end{IPA} & \begin{IPA}\textbf{\textltailn}i\end{IPA} & \begin{IPA}\textbf{\textltailn}i\end{IPA} & \begin{IPA}\textbf{\textltailn}i\end{IPA} & \begin{IPA}\textbf{n}i\end{IPA} & \begin{IPA}\textbf{\ng}idEm\end{IPA} & no \\
\begin{IPA}/a/\end{IPA} before nasal & \begin{IPA}p\textbf{1}n\end{IPA} & \begin{IPA}p\textbf{a}n\end{IPA} & \begin{IPA}p\textbf{a}n\end{IPA} & \begin{IPA}p\textbf{a}n\end{IPA} & \begin{IPA}p\textbf{a}n\end{IPA} & \begin{IPA}p\textbf{a}n\end{IPA} & \begin{IPA}p\textbf{a}n\end{IPA} & yes \\
\begin{IPA}/a/\end{IPA} blocked syllable & \begin{IPA}p\textbf{a}p\end{IPA} & \begin{IPA}p\textbf{a}p\end{IPA} & \begin{IPA}p\textbf{a}p\end{IPA} & \begin{IPA}p\textbf{a}p\end{IPA} & \begin{IPA}p\textbf{a}p\end{IPA} & \begin{IPA}p\textbf{a}p\end{IPA} & \begin{IPA}p\textbf{a}p\end{IPA} & yes \\
\begin{IPA}/i/\end{IPA} & \begin{IPA}p\textbf{i}\end{IPA} & \begin{IPA}p\textbf{i}\end{IPA} & \begin{IPA}p\textbf{i}\end{IPA} & \begin{IPA}p\textbf{i}\end{IPA} & \begin{IPA}p\textbf{i}\end{IPA} & \begin{IPA}p\textbf{i}\end{IPA} & \begin{IPA}p\textbf{i}\end{IPA} & yes \\
\begin{IPA}/u/\end{IPA} & \begin{IPA}p\textbf{u}\end{IPA} & \begin{IPA}p\textbf{y}\end{IPA} & \begin{IPA}p\textbf{u}\end{IPA} & \begin{IPA}p\textbf{u}\end{IPA} & \begin{IPA}p\textbf{u}\end{IPA} & \begin{IPA}p\textbf{u}\end{IPA} & \begin{IPA}p\textbf{u}\end{IPA} & yes \\
\begin{IPA}/I/\end{IPA} blocked syllable & \begin{IPA}p\textbf{e}p\end{IPA} & \begin{IPA}p\textbf{e}p\end{IPA} & \begin{IPA}p\textbf{e}p\end{IPA} & \begin{IPA}p\textbf{e}p\end{IPA} & \begin{IPA}p\textbf{e}p\end{IPA} & \begin{IPA}p\textbf{I}p\end{IPA} & \begin{IPA}p\textbf{I}p\end{IPA} & yes \\
\begin{IPA}/U/\end{IPA} blocked syllable & \begin{IPA}p\textbf{u}p\end{IPA} & \begin{IPA}p\textbf{u}p\end{IPA} & \begin{IPA}p\textbf{o}p\end{IPA} & \begin{IPA}p\textbf{o}p\end{IPA} & \begin{IPA}p\textbf{o}p\end{IPA} & \begin{IPA}p\textbf{U}p\end{IPA} & \begin{IPA}p\textbf{U}p\end{IPA} & yes \\
\begin{IPA}/O/\end{IPA} blocked syllable & \begin{IPA}p\textbf{o}p\end{IPA} & \begin{IPA}p\textbf{O}p\end{IPA} & \begin{IPA}p\textbf{O}p\end{IPA} & \begin{IPA}p\textbf{we}p\end{IPA} & \begin{IPA}p\textbf{O}p\end{IPA} & \begin{IPA}p\textbf{O}p\end{IPA} & \begin{IPA}p\textbf{O}p\end{IPA} & yes \\
\begin{IPA}/k/\end{IPA} before front vowels & \begin{IPA}\textbf{tS}i\end{IPA} & \begin{IPA}\textbf{s}i\end{IPA} & \begin{IPA}\textbf{tS}i\end{IPA} & \begin{IPA}\textbf{T}i\end{IPA} & \begin{IPA}\textbf{s}i\end{IPA} & \begin{IPA}\textbf{k}i\end{IPA} & \begin{IPA}\textbf{k}i\end{IPA} & yes \\
\begin{IPA}/sk/\end{IPA} before front vowels & \begin{IPA}\textbf{St}i\end{IPA} & \begin{IPA}\textbf{s}i\end{IPA} & \begin{IPA}\textbf{S}i\end{IPA} & \begin{IPA}\textbf{T}i\end{IPA} & \begin{IPA}\textbf{S}i\end{IPA} & \begin{IPA}\textbf{sk}i\end{IPA} & \begin{IPA}\textbf{sk}i\end{IPA} & yes \\
\begin{IPA}/kt/\end{IPA} medially, elsewhere & \begin{IPA}a\textbf{pt}a\end{IPA} & \begin{IPA}a\textbf{t}a\end{IPA} & \begin{IPA}a\textbf{tt}a\end{IPA} & \begin{IPA}a\textbf{tS}a\end{IPA} & \begin{IPA}a\textbf{t}a\end{IPA} & \begin{IPA}a\textbf{kt}a\end{IPA} & \begin{IPA}a\textbf{kt}am\end{IPA} & yes \\
\begin{IPA}/aU/\end{IPA} & \begin{IPA}p\textbf{au}\end{IPA} & \begin{IPA}p\textbf{O}\end{IPA} & \begin{IPA}p\textbf{O}\end{IPA} & \begin{IPA}p\textbf{o}\end{IPA} & \begin{IPA}p\textbf{o}\end{IPA} & \begin{IPA}p\textbf{aU}\end{IPA} & \begin{IPA}p\textbf{aU}m\end{IPA} & yes \\
\begin{IPA}/pl/\end{IPA} word initial & \begin{IPA}\textbf{pl}a\end{IPA} & \begin{IPA}\textbf{pl}a\end{IPA} & \begin{IPA}\textbf{pj}a\end{IPA} & \begin{IPA}\textbf{L}a\end{IPA} & \begin{IPA}\textbf{S}a\end{IPA} & \begin{IPA}\textbf{pl}a\end{IPA} & \begin{IPA}\textbf{pl}am\end{IPA} & yes \\
\begin{IPA}/a/\end{IPA} free syllable & \begin{IPA}p\textbf{a}\end{IPA} & \begin{IPA}p\textbf{a}\end{IPA} & \begin{IPA}p\textbf{a}\end{IPA} & \begin{IPA}p\textbf{a}\end{IPA} & \begin{IPA}p\textbf{a}\end{IPA} & \begin{IPA}p\textbf{a}\end{IPA} & \begin{IPA}p\textbf{a}m\end{IPA} & yes \\
\begin{IPA}/E/\end{IPA} free syllable & \begin{IPA}p\textbf{je}\end{IPA} & \begin{IPA}p\textbf{je}\end{IPA} & \begin{IPA}p\textbf{je}\end{IPA} & \begin{IPA}p\textbf{je}\end{IPA} & \begin{IPA}p\textbf{E}\end{IPA} & \begin{IPA}p\textbf{E}\end{IPA} & \begin{IPA}p\textbf{E}m\end{IPA} & yes \\
\begin{IPA}/w/\end{IPA} & \begin{IPA}\textbf{v}a\end{IPA} & \begin{IPA}\textbf{v}a\end{IPA} & \begin{IPA}\textbf{v}a\end{IPA} & \begin{IPA}\textbf{b}a\end{IPA} & \begin{IPA}\textbf{v}a\end{IPA} & \begin{IPA}\textbf{w}a\end{IPA} & \begin{IPA}\textbf{w}am\end{IPA} & yes \\
\begin{IPA}/b/\end{IPA} word initial & \begin{IPA}\textbf{b}a\end{IPA} & \begin{IPA}\textbf{b}a\end{IPA} & \begin{IPA}\textbf{b}a\end{IPA} & \begin{IPA}\textbf{b}a\end{IPA} & \begin{IPA}\textbf{b}a\end{IPA} & \begin{IPA}\textbf{b}a\end{IPA} & \begin{IPA}\textbf{b}am\end{IPA} & yes \\
\begin{IPA}/j/\end{IPA} word initial & \begin{IPA}\textbf{Z}a\end{IPA} & \begin{IPA}\textbf{Z}a\end{IPA} & \begin{IPA}\textbf{dZ}a\end{IPA} & \begin{IPA}\textbf{x}a\end{IPA} & \begin{IPA}\textbf{Z}a\end{IPA} & \begin{IPA}\textbf{j}a\end{IPA} & \begin{IPA}\textbf{j}am\end{IPA} & yes \\
\begin{IPA}/f/\end{IPA} word initial & \begin{IPA}\textbf{f}a\end{IPA} & \begin{IPA}\textbf{f}a\end{IPA} & \begin{IPA}\textbf{f}a\end{IPA} & \begin{IPA}a\end{IPA} & \begin{IPA}\textbf{f}a\end{IPA} & \begin{IPA}\textbf{f}a\end{IPA} & \begin{IPA}\textbf{f}am\end{IPA} & yes \\
\begin{IPA}/f/\end{IPA} elsewhere & \begin{IPA}a\textbf{f}a\end{IPA} & \begin{IPA}a\textbf{f}a\end{IPA} & \begin{IPA}a\textbf{f}a\end{IPA} & \begin{IPA}a\textbf{f}a\end{IPA} & \begin{IPA}a\textbf{f}a\end{IPA} & \begin{IPA}a\textbf{f}a\end{IPA} & \begin{IPA}a\textbf{f}fam\end{IPA} & yes \\
\begin{IPA}/U/\end{IPA} free syllable & \begin{IPA}p\textbf{u}\end{IPA} & \begin{IPA}p\textbf{\o}\end{IPA} & \begin{IPA}p\textbf{o}\end{IPA} & \begin{IPA}p\textbf{o}\end{IPA} & \begin{IPA}p\textbf{o}\end{IPA} & \begin{IPA}p\textbf{U}\end{IPA} & \begin{IPA}p\textbf{U}pUm\end{IPA} & yes \\
\begin{IPA}/O/\end{IPA} free syllable & \begin{IPA}p\textbf{o}\end{IPA} & \begin{IPA}p\textbf{\o}\end{IPA} & \begin{IPA}p\textbf{wO}\end{IPA} & \begin{IPA}p\textbf{we}\end{IPA} & \begin{IPA}p\textbf{O}\end{IPA} & \begin{IPA}p\textbf{O}\end{IPA} & \begin{IPA}p\textbf{O}dEm\end{IPA} & yes \\
\begin{IPA}/l/\end{IPA} before front vowels & \begin{IPA}\textbf{j}i\end{IPA} & \begin{IPA}\textbf{j}i\end{IPA} & \begin{IPA}\textbf{L}i\end{IPA} & \begin{IPA}\textbf{x}i\end{IPA} & \begin{IPA}\textbf{L}i\end{IPA} & \begin{IPA}\textbf{l}i\end{IPA} & \begin{IPA}gI\textbf{l}Um\end{IPA} & yes \\
\end{tabular}}
\end{scalebox}

            \caption{the set of test phonemes used to evaluate the model's generalizations. Each row represents a distinct rule of phonetic change, which focuses on a single phoneme. The phoneme in question is bolded, and other consonants / vowels are added to simulate the phonological environment of the rule. The added consonants / vowels were chosen because they did not affect the evolution of the examined phonemes from Latin to the Romance languages. ``correct" signifies whether the network's prediction were correct.}
    \label{appendix:table-test-rules}

\end{table*}

To what extent did the model learn known rules of phonetic change?

The evolution of the Romance languages is well studied and linguists documented the set of phonological transformations that underwent between Latin and its daughter languages.
%In order to test which rules of change the model truly learned, we ran a controlled experiment which evaluated how well the network performed on specific historical changes. 
We collected 33 of these phonological change rules, and used them to create a ``synthetic'' test set, containing syllable examples each focusing on a different phonological change. An example of a row in this dataset, corresponding to the rule of change of Latin [j] at word initial, is:\\

$x=$\begin{IPA}Za\end{IPA}$^{\textsc{Rm}}$,
\begin{IPA}Za\end{IPA}$^{\textsc{Fr}}$,
\begin{IPA}dZa\end{IPA}$^\textsc{It}$,
\begin{IPA}xa\end{IPA}$^\textsc{Sp}$,
\begin{IPA}Za\end{IPA}$^\textsc{Pt}$

$y=$\begin{IPA}ja\end{IPA}

Since the model was trained on complete words, isolated syllables tended to be unnatural for the network, and the output often contained additional consonants (usually morphological endings). When evaluating the model output we focus on the specific phonemes involved in the phonological change, and we ignore additional phonological material. 

\paragraph{Results}
The complete list of synthetic examples and predictions is available at Table \ref{appendix:table-test-rules}. The network correctly predicted 22 out of the 33 phonological rules (66.67\% of the changes). The results are compatible with the results of the main reconstruction experiment: In both experiments, the network correctly reconstructed phonemes retained with little or no changes in all languages (e.g. \begin{IPA}[a]\end{IPA} in different -phonological environments). Another class of phonemes correctly reconstructed in both cases are those which changed in a predictable way in each one of the daughter languages. Thus, \begin{IPA}[w]\end{IPA} was correctly reconstructed since it predictably changed to \begin{IPA}[v]\end{IPA} in all the daughter languages (apart from Spanish, which merged it with \begin{IPA}[b]\end{IPA}). Phonemes that tended to change differently, but consistently, were also faithfully recovered: even though Latin \begin{IPA}[k]\end{IPA} tended to change differently depending on the daughter language (\begin{IPA}[s]\end{IPA} in French and Portuguese, \begin{IPA}[T]\end{IPA} in Spanish and \begin{IPA}[tS]\end{IPA} in Italian and Romanian), it was reconstructed correctly because of the consistence of the change in each daughter language.
The phonemes wrongly reconstructed tended to be those whose phonological change was ``opaque''. The ``opaqueness'' of their change can be ascribed to the fact that they were neutralized in the daughter languages, making it impossible to recover them without additional information. Relevant to this case are mostly vowels and diphthongs, as Latin \begin{IPA}[e]\end{IPA} and \begin{IPA}[I]\end{IPA}, which both became \begin{IPA}[e]\end{IPA} in all the different daughter languages (with variants influenced by the phonological environment).

\subsection{Learnt phoneme representations}
\label{clustering}

\begin{figure}
    \includegraphics[width=0.5\textwidth]{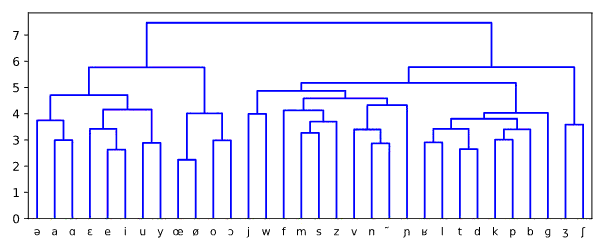}
    %\vspace{-4em}
        \caption{Hierarchical clustering of French phoneme embeddings}
    \label{plot:clust-fr}
\end{figure}

%The phonological diachronic change is not arbitrary, and it is tightly influenced by the human articulatory apparatus. Hence, the same processes occur repeatedly in the world languages \cite{Anttila}.
Does training on proto-word reconstructions implicitly encourage the model to acquire phonologically-meaningful representations? We visualize the representation learned by network on the phonetic task by performing hierarchical clustering on the characters embedding vectors using the sklearn \cite{sklearn} implementation of Ward variance minimization algorithm \cite{ward1963hierarchical}.

Here we will briefly discuss the learned French phoneme representations (Figure \ref{plot:clust-fr}). For all other languages, see  appendix \S \ref{appendix:clust-all}.  As can be seen, the primary division that the network performs is between  vowels and consonants, displayed on two different branches of the tree. On a lower level other phonologically motivated groupings are found: the network tends to place under the same node pairs of voiced and unvoiced consonants (as \begin{IPA}[S]\end{IPA} and \begin{IPA}[Z]\end{IPA}, \begin{IPA}[d]\end{IPA} and \begin{IPA}[t]\end{IPA}), allophones (\begin{IPA}[\oe]\end{IPA} and \begin{IPA}[\o]\end{IPA}) or phonemes of the same category (as the glides \begin{IPA}[j]\end{IPA} and \begin{IPA}[w]\end{IPA}). To conclude, the results demonstrate the learning of a phonologically meaningful taxonomy of phonemes, without explicit supervision. 

\begin{figure}
    \includegraphics[scale=0.45, left]{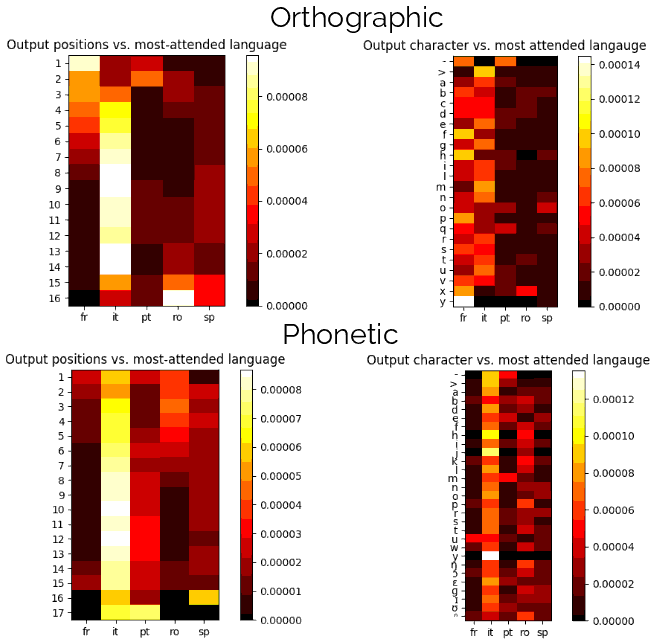}
        \caption{position in output vs. most attended language (left) and output letter vs. most attended langauge (right) for the orthographic (upper) and phonetic (lower) tasks.}
    \label{attention-plots}
\end{figure}

\subsection{Attention analysis}
\label{section-attention}

%Within a given language family, different languages can change to a varying extent compared with their common proto-language: while most North Germanic languages (as Norwegian and Swedish) have greatly reduced levels of morphological inflection, Icelandic retains a considerably more conservative and synthetic morphology, closer to the one of the common ancestor of all North Germanic languages \cite{einarsson1945icelandic}. 
Since different languages can diverge to a varying extent from their proto-language, we hypothesize that the 5 daughter languages we use in this work would be of different importance for the model. To test this hypothesis, we inspect the learned attention weights. We focus on the most attended input character at each time step (the character having the largest attention weight) and count the number of times each of the 5 input languages is the most attended language, as a function of the location in the output and of the identity of the Latin character produced in that time step. We normalize the count with respect to time step, letter frequency and language frequency in the corpus.

\paragraph{Results} The results for the phonetic and orthographic tasks are presented in Figure \ref{attention-plots}. In both cases, Italian is the most attended language. There are some differences between the settings, however. For the orthographic task, the network focuses noticeably more on French than in the phonetic task. This tendency can be attributed to the very conservative orthography of French, that masks the phonological innovations that occurred in the language. Indeed, the network focuses exclusively on French for the reconstruction of the characters $\textless$h$\textgreater$ and $\textless$y$\textgreater$, which are consistently represented only in French orthography, disappearing from the written form of the other Romance languages. The comparison to the attention of the phonetic dataset shows that the network tends to actually ignore French, favoring other sources instead. Similarly, in the orthographic dataset, French is favored in the initial positions, a tendency that disappears in the phonetic dataset. 
Finally, an interesting trend in the phonetic dataset is a tendency to attend to Romanian at the initial positions and to Portuguese at later ones.

\section{Conclusions}

In this work, we introduce a new dataset for the task of proto-word reconstruction in the Romance language family, and used it to evaluate the ability of neural networks to capture the regularities of historic language change. We have shown that neural methods outperform previously suggested models for this task.  Analysis of the linguistic generalizations the model acquires during training demonstrated that the mistakes are related to the complexity of the phonetic change. A controlled experiment on a set of rules for phonetic alternations between Latin and its daughter languages demonstrated the model internalizes some of the systematic processes that Latin had undergone during the evolution of the Romance languages. Visualizing the learned phoneme-embedding vectors has revealed a hierarchical division of phonemes that reflects phonological realities, and inspection of attention patterns demonstrated the model attributes different importance to different languages, in a position-dependent manner. 

While the task examined in this paper is commonly called "proto-word reconstruction", in practice the task the model faces is considerably less challenging than the work of historical linguists, as the model is trained in a supervised setting. A future line of work we suggest is applying neural models for the end task of proto word reconstruction, without relying on cognates lists, in a way that would more naturally model the historical linguistic methodology. 

\section*{Acknowledgements}
We thank Arya McCarthy for pointing out to relevant references.
This project received funding from the Europoean Research Council (ERC) under the Europoean Union's Horizon 2020 research and innovation programme, grant agreement No. 802774 (iEXTRACT).

\bibliographystyle{acl_natbib}
\bibliography{latin}

\begin{thebibliography}{37}
\expandafter\ifx\csname natexlab\endcsname\relax\def\natexlab#1{#1}\fi

\bibitem[{Alkire and Rosen(2010)}]{HistoricalIntroduction}
Ti~Alkire and Carol Rosen. 2010.
\newblock \emph{Romance languages: A historical introduction}.
\newblock Cambridge University Press.

\bibitem[{Allen and Allen(1989)}]{LtOrtho}
W~Sidney Allen and William~Sidney Allen. 1989.
\newblock \emph{Vox latina}.
\newblock Cambridge University Press.

\bibitem[{Anttila(1989)}]{Anttila}
Raimo Anttila. 1989.
\newblock \emph{Historical and comparative linguistics}, volume~6.
\newblock John Benjamins Publishing.

\bibitem[{Bahdanau et~al.(2015)Bahdanau, Cho, and
  Bengio}]{encoder-decoder-attention}
Dzmitry Bahdanau, Kyunghyun Cho, and Yoshua Bengio. 2015.
\newblock \href {http://arxiv.org/abs/1409.0473} {Neural machine translation by
  jointly learning to align and translate}.
\newblock In \emph{3rd International Conference on Learning Representations,
  {ICLR} 2015, San Diego, CA, USA, May 7-9, 2015, Conference Track
  Proceedings}.

\bibitem[{Bouchard-C{\^o}t{\'e} et~al.(2009)Bouchard-C{\^o}t{\'e}, Griffiths,
  and Klein}]{bouchard2009improved}
Alexandre Bouchard-C{\^o}t{\'e}, Thomas~L Griffiths, and Dan Klein. 2009.
\newblock Improved reconstruction of protolanguage word forms.
\newblock In \emph{Proceedings of human language technologies: The 2009 annual
  conference of the north american chapter of the association for computational
  linguistics}, pages 65--73. Association for Computational Linguistics.

\bibitem[{Bouchard-C{\^o}t{\'e} et~al.(2013)Bouchard-C{\^o}t{\'e}, Hall,
  Griffiths, and Klein}]{bouchard2013automated}
Alexandre Bouchard-C{\^o}t{\'e}, David Hall, Thomas~L Griffiths, and Dan Klein.
  2013.
\newblock Automated reconstruction of ancient languages using probabilistic
  models of sound change.
\newblock \emph{Proceedings of the National Academy of Sciences},
  110(11):4224--4229.

\bibitem[{Bouchard{-}C{\^{o}}t{\'{e}} et~al.(2007)Bouchard{-}C{\^{o}}t{\'{e}},
  Liang, Griffiths, and Klein}]{Bouchard2007}
Alexandre Bouchard{-}C{\^{o}}t{\'{e}}, Percy Liang, Thomas~L. Griffiths, and
  Dan Klein. 2007.
\newblock \href {http://www.aclweb.org/anthology/D07-1093} {A probabilistic
  approach to diachronic phonology}.
\newblock In \emph{EMNLP-CoNLL 2007, Proceedings of the 2007 Joint Conference
  on Empirical Methods in Natural Language Processing and Computational Natural
  Language Learning, June 28-30, 2007, Prague, Czech Republic}, pages 887--896.

\bibitem[{Boyd-Bowman(1980)}]{LatinToRomance}
Peter Boyd-Bowman. 1980.
\newblock \emph{From Latin to Romance in sound charts}.
\newblock Georgetown University Press.

\bibitem[{Cho et~al.(2014)Cho, van Merrienboer, Bahdanau, and
  Bengio}]{bengio-NMT}
Kyunghyun Cho, Bart van Merrienboer, Dzmitry Bahdanau, and Yoshua Bengio. 2014.
\newblock \href {http://aclweb.org/anthology/W/W14/W14-4012.pdf} {On the
  properties of neural machine translation: Encoder-decoder approaches}.
\newblock In \emph{Proceedings of SSST@EMNLP 2014, Eighth Workshop on Syntax,
  Semantics and Structure in Statistical Translation, Doha, Qatar, 25 October
  2014}, pages 103--111.

\bibitem[{Ciobanu and Dinu(2014{\natexlab{a}})}]{ciobanu2014automatic}
Alina~Maria Ciobanu and Liviu~P Dinu. 2014{\natexlab{a}}.
\newblock Automatic detection of cognates using orthographic alignment.
\newblock In \emph{Proceedings of the 52nd Annual Meeting of the Association
  for Computational Linguistics (Volume 2: Short Papers)}, volume~2, pages
  99--105.

\bibitem[{Ciobanu and Dinu(2014{\natexlab{b}})}]{ciobanu-dataset}
Alina~Maria Ciobanu and Liviu~P Dinu. 2014{\natexlab{b}}.
\newblock Building a dataset of multilingual cognates for the romanian lexicon.
\newblock In \emph{Proceedings of the 9th International Conference on Language
  Resources and Evaluation, LREC}, pages 1038--1043.

\bibitem[{Ciobanu and Dinu(2018)}]{ciobanu18}
Alina~Maria Ciobanu and Liviu~P Dinu. 2018.
\newblock Ab initio: Automatic latin proto-word reconstruction.
\newblock In \emph{Proceedings of the 27th International Conference on
  Computational Linguistics}, pages 1604--1614.

\bibitem[{Clegg and Fails(2017)}]{SpOrtho}
J~Halvor Clegg and Willis~C Fails. 2017.
\newblock \emph{Manual de fon{\'e}tica y fonolog{\'\i}a espa{\~n}olas}.
\newblock Routledge.

\bibitem[{Debove and Rey(2000)}]{FrOrtho}
Josette~Rey Debove and Alain Rey. 2000.
\newblock \emph{Le Nouveau Petit Robert: Dictionnaire alphabetique et
  analogique de la langue francaise}.
\newblock Dictionnaires Le Robert.

\bibitem[{DIEZ and Donkin(1864)}]{etymologicalDict}
Friedrich~Christian DIEZ and TC~Donkin. 1864.
\newblock \emph{An Etymological Dictionary of the Romance Languages; chiefly
  from the German of F. Diez. By TC Donkin}.
\newblock Williams and Norgate.

\bibitem[{Ernst(2003)}]{ernst2003romanische}
Gerhard Ernst. 2003.
\newblock \emph{Romanische Sprachgeschichte: Ein Internationales Handbuch Zur
  Geschichte Der Romanischen Sprachen Und Ihrer Erforschung/Manuel
  International Sur L'Histoire Et L'Etude Linguistique DES Langues Romanes}.
\newblock Mouton de Gruyter.

\bibitem[{Gaffiot and Flobert(1934)}]{Gaffiot}
F{\'e}lix Gaffiot and Pierre Flobert. 1934.
\newblock \emph{Dictionnaire latin-fran{\c{c}}ais}.
\newblock Hachette Paris.

\bibitem[{Hall(1944)}]{ItnOrtho}
Robert~A Hall. 1944.
\newblock Italian phonemes and orthography.
\newblock \emph{Italica}, 21(2):72--82.

\bibitem[{Harris and Vincent(2003)}]{RomanceLanguages}
Martin Harris and Nigel Vincent. 2003.
\newblock \emph{The romance languages}.
\newblock Routledge.

\bibitem[{Holtus et~al.(1989)Holtus, Metzeltin, and Schmitt}]{holtus1989lrl}
G{\"u}nter Holtus, Michael Metzeltin, and Christian Schmitt. 1989.
\newblock \emph{LRL}, volume~3.
\newblock M. Niemeyer.

\bibitem[{Inkpen et~al.(2005)Inkpen, Frunza, and Kondrak}]{inkpen2005automatic}
Diana Inkpen, Oana Frunza, and Grzegorz Kondrak. 2005.
\newblock Automatic identification of cognates and false friends in french and
  english.
\newblock In \emph{Proceedings of the International Conference Recent Advances
  in Natural Language Processing}, volume~9, pages 251--257.

\bibitem[{Kondrak(2001)}]{kondrak2001identifying}
Grzegorz Kondrak. 2001.
\newblock Identifying cognates by phonetic and semantic similarity.
\newblock In \emph{Proceedings of the second meeting of the North American
  Chapter of the Association for Computational Linguistics on Language
  technologies}, pages 1--8. Association for Computational Linguistics.

\bibitem[{Ledgeway and Maiden(2016)}]{ledgeway2016oxford}
Adam Ledgeway and Martin Maiden. 2016.
\newblock \emph{The Oxford guide to the Romance languages}, volume~1.
\newblock Oxford University Press.

\bibitem[{Lewis and Short(1879)}]{Charlton}
Charlton~T. Lewis and Charles Short. 1879.
\newblock A latin dictionary.
\newblock \emph{Perseus Digital Library}.

\bibitem[{Lewis et~al.(2020)Lewis, Wu, McCarthy, and
  Yarowsky}]{DBLP:conf/coling/LewisWMY20}
Dylan Lewis, Winston Wu, Arya~D. McCarthy, and David Yarowsky. 2020.
\newblock \href {https://doi.org/10.18653/v1/2020.coling-main.387} {Neural
  transduction for multilingual lexical translation}.
\newblock In \emph{Proceedings of the 28th International Conference on
  Computational Linguistics, {COLING} 2020, Barcelona, Spain (Online), December
  8-13, 2020}, pages 4373--4384. International Committee on Computational
  Linguistics.

\bibitem[{List et~al.(2016)List, Lopez, and Bapteste}]{list2016using}
Johann-Mattis List, Philippe Lopez, and Eric Bapteste. 2016.
\newblock Using sequence similarity networks to identify partial cognates in
  multilingual wordlists.
\newblock In \emph{Proceedings of the 54th Annual Meeting of the Association
  for Computational Linguistics (Volume 2: Short Papers)}, volume~2, pages
  599--605.

\bibitem[{Mann and Yarowsky(2001)}]{mann-distance-induction}
Gideon~S. Mann and David Yarowsky. 2001.
\newblock \href {http://aclweb.org/anthology/N/N01/N01-1020.pdf} {Multipath
  translation lexicon induction via bridge languages}.
\newblock In \emph{Language Technologies 2001: The Second Meeting of the North
  American Chapter of the Association for Computational Linguistics, {NAACL}
  2001, Pittsburgh, PA, USA, June 2-7, 2001}.

\bibitem[{Mateus and d'Andrade(2000)}]{PtOrtho}
Maria~Helena Mateus and Ernesto d'Andrade. 2000.
\newblock \emph{The phonology of Portuguese}.
\newblock OUP Oxford.

\bibitem[{Millar(2013)}]{Trask}
Robert~McColl Millar. 2013.
\newblock \emph{Trask's historical linguistics}.
\newblock Routledge.

\bibitem[{Mulloni and Pekar(2006)}]{mulloni2006automatic}
Andrea Mulloni and Viktor Pekar. 2006.
\newblock Automatic detection of orthographics cues for cognate recognition.
\newblock In \emph{LREC}, pages 2387--2390.

\bibitem[{Nishimura et~al.(2020)Nishimura, Sudoh, Neubig, and
  Nakamura}]{DBLP:journals/taslp/NishimuraSNN20}
Yuta Nishimura, Katsuhito Sudoh, Graham Neubig, and Satoshi Nakamura. 2020.
\newblock \href {https://doi.org/10.1109/TASLP.2019.2959224} {Multi-source
  neural machine translation with missing data}.
\newblock \emph{{IEEE} {ACM} Trans. Audio Speech Lang. Process.}, 28:569--580.

\bibitem[{Pedregosa et~al.(2011)Pedregosa, Varoquaux, Gramfort, Michel,
  Thirion, Grisel, Blondel, Prettenhofer, Weiss, Dubourg et~al.}]{sklearn}
Fabian Pedregosa, Ga{\"e}l Varoquaux, Alexandre Gramfort, Vincent Michel,
  Bertrand Thirion, Olivier Grisel, Mathieu Blondel, Peter Prettenhofer, Ron
  Weiss, Vincent Dubourg, et~al. 2011.
\newblock Scikit-learn: Machine learning in python.
\newblock \emph{Journal of machine learning research}, 12(Oct):2825--2830.

\bibitem[{Rama et~al.(2018)Rama, List, Wahle, and
  J{\"{a}}ger}]{rama-cognates-eval}
Taraka Rama, Johann{-}Mattis List, Johannes Wahle, and Gerhard J{\"{a}}ger.
  2018.
\newblock \href {https://aclanthology.info/papers/N18-2063/n18-2063} {Are
  automatic methods for cognate detection good enough for phylogenetic
  reconstruction in historical linguistics?}
\newblock In \emph{Proceedings of the 2018 Conference of the North American
  Chapter of the Association for Computational Linguistics: Human Language
  Technologies, NAACL-HLT, New Orleans, Louisiana, USA, June 1-6, 2018, Volume
  2 (Short Papers)}, pages 393--400.

\bibitem[{Sarlin(2014)}]{RmOrtho}
Mika Sarlin. 2014.
\newblock \emph{Romanian grammar}.
\newblock BoD-Books on Demand.

\bibitem[{Ward~Jr(1963)}]{ward1963hierarchical}
Joe~H Ward~Jr. 1963.
\newblock Hierarchical grouping to optimize an objective function.
\newblock \emph{Journal of the American statistical association},
  58(301):236--244.

\bibitem[{Wu et~al.(2020)Wu, Nicolai, and Yarowsky}]{DBLP:conf/lrec/WuNY20}
Winston Wu, Garrett Nicolai, and David Yarowsky. 2020.
\newblock \href {https://www.aclweb.org/anthology/2020.lrec-1.519/}
  {Multilingual dictionary based construction of core vocabulary}.
\newblock In \emph{Proceedings of The 12th Language Resources and Evaluation
  Conference, {LREC} 2020, Marseille, France, May 11-16, 2020}, pages
  4211--4217. European Language Resources Association.

\bibitem[{Wu and Yarowsky(2018)}]{DBLP:conf/lrec/WuY18}
Winston Wu and David Yarowsky. 2018.
\newblock \href
  {http://www.lrec-conf.org/proceedings/lrec2018/summaries/934.html} {Creating
  large-scale multilingual cognate tables}.
\newblock In \emph{Proceedings of the Eleventh International Conference on
  Language Resources and Evaluation, {LREC} 2018, Miyazaki, Japan, May 7-12,
  2018}. European Language Resources Association {(ELRA)}.

\end{thebibliography}

\clearpage

\appendix

\section{Appendix}

\subsection{Dataset Creation}
\label{appendix:dataset-creation}

In order to perform the reconstruction task, we required a large dataset of cognates and their proto-words, in both orthographic and phonetic (IPA) forms.
%phonetic versions of the datase

Despite growing interest in recent years, high-quality digital resources for the tasks of proto-word reconstruction and cognates detection are scarce. Our departure point is the dataset provided by \citet{ciobanu-dataset}, which, to the best of our knowledge, is the most extensive dataset for proto-word reconstruction of a well-attested proto-language. The dataset contains 3,218 complete cognate sets in five Romance languages (Spanish, Italian, Portuguese, French, Romanian) together with their Latin etymological ancestor. Although being a valuable resource, this dataset was constructed via automatic method of cognate extraction, and a comparison with references on the development of Romance languages \cite{LatinToRomance, HistoricalIntroduction} reveals some problems, such as false cognates, truncated forms, non-existent words and mismatch between the part of speech of the cognates and the ancestor. %\ym{is it clear? what I wanted to express is that several times one of the cognates was from a different part of speech comparing to the Latin original word}. 
Another salient problem of the dataset regarded the grammatical case of Latin nouns: Romance languages derived their words from the accusative Latin case \cite{RomanceLanguages}, while in the dataset Latin words were displayed in the nominative case, an inconsistency making the reconstruction inherently more challenging.

Lastly, as neural models often requires large amounts of training data, we aimed to expand the dataset. 
We thus created a cleaned and extended dataset by Wiktionary scrapping, followed by manual validation and cleansing.

\paragraph{Wikitionary scraping}
We augment the existing dataset with a freely available resource: Wiktionary. Wiktionary entries for Latin words usually contain inflection tables, and often list the descendants in Romance languages; these descendents are, by definition, cognates. 
We scraped all Latin entries from Wiktionary, and extracted the forms of the daughter languages (available in the ``Descendants'' section). This resulted in 5,598 additional comparative entries, for a total of 22,361 new individual words. 
Contrary to the previous dataset, the Wiktionary-derived cognates are not based on automatic alignment between translations, but rather on direct human annotation. On the other hand, the Wiktionary-based entries are often incomplete, and include cognates in only a subset of the daughter languages.

\paragraph{Form normalization} Using the Wiktionary-provided inflection tables, we decline the Latin nouns to the accusative case, and conjugate verbs to the infinitive form. We do this both to the Wiktionary-based entries and to the ones in the original dataset. We selected a sample of around 100 Latin words to check the accuracy of the automatic conjugation, against \citet{Gaffiot}, finding them all correct. Finally, Latin words in the \citet{ciobanu-dataset} dataset for which we did not find a Wiktionary entry were conjugated ``manually'' by consulting \cite{Charlton,  Gaffiot}.

\paragraph{Manual verification and cleaning} After the collection of the Wikitionary dataset, we went manually through all the Latin words contained in \citet{ciobanu-dataset}, checking them against \citet{Charlton, Gaffiot}. Additionally, we went over the some suspicious-looking words from the daughter language and verified them against \citet{etymologicalDict} to ensure their etymological relatedness with the Latin source, fixing if necessary.This sort of fix was not performed systematically, but we did fix or remove around 170 words.

Finally, we sample 300 entries from the original \cite{ciobanu-dataset} dataset prior to cleaning and 300 words from our cleaned and unified version of the dataset, and manually verified them. We find 43 mistakes in the original dataset and only 4 in our version, indicating that, while still not perfect, it is of substantial higher quality.

\paragraph{IPA transcription} 
To obtain the phonetic transcriptions into IPA, we utilized the transcription module of the eSpeak library, which offers transcriptions for all languages in our dataset, including Latin. While a human transcription would be preferable, a manual evaluation of 200 of the resulting transcriptions 
by comparing them 
against several sources \cite{LtOrtho, ItnOrtho,FrOrtho,SpOrtho,PtOrtho,RmOrtho} show high accuracy: all the 200 words were correct, except for minor systematic changes which we fixed globally to better suit the transcription to phonological conventions. Specifically, we deleted the vowel symbols $\textless$\begin{IPA}U\end{IPA}$\textgreater$ and $\textless$\begin{IPA}I\end{IPA}$\textgreater$ in Italian and Romanian, which resulted to be alien to those languages, changed the sequence $\textless$\begin{IPA}RR\end{IPA}$\textgreater$ to $\textless$\begin{IPA}r\end{IPA}$\textgreater$ in Spanish, and regularized the Portuguese transcriptions, which showed some phonological traits of Brazilian Portuguese.

\paragraph{Final dataset} The resulting dataset, used for all experiments in this work, contains 8,799 entries. The dataset was randomly splitted into train, evaluation and test sets, with 7,038 examples (80\%) used for training, 703 (8\%) for evaluation and 1,055 (12\%) for testing. 

Overall, the dataset contains 41,563 distinct words across the different languages (for a total of 83,126 words counting both the orthographic and the phonetic datasets), with 7,384 Italian words, 7,183 Spanish words, 6,806 Portuguese words, 6,505 French words and 4,886 Romanian words. As vowel lengths were found to be difficult to recover, we created the following variations of the dataset: with and without vowel length (for both the orthographic and phonetic datasets), and without a contrast (for the phonetic dataset).

\clearpage
\subsection{Phoneme representations}
\label{appendix:clustering}

\begin{figure}[H]
    \includegraphics[scale=0.31]{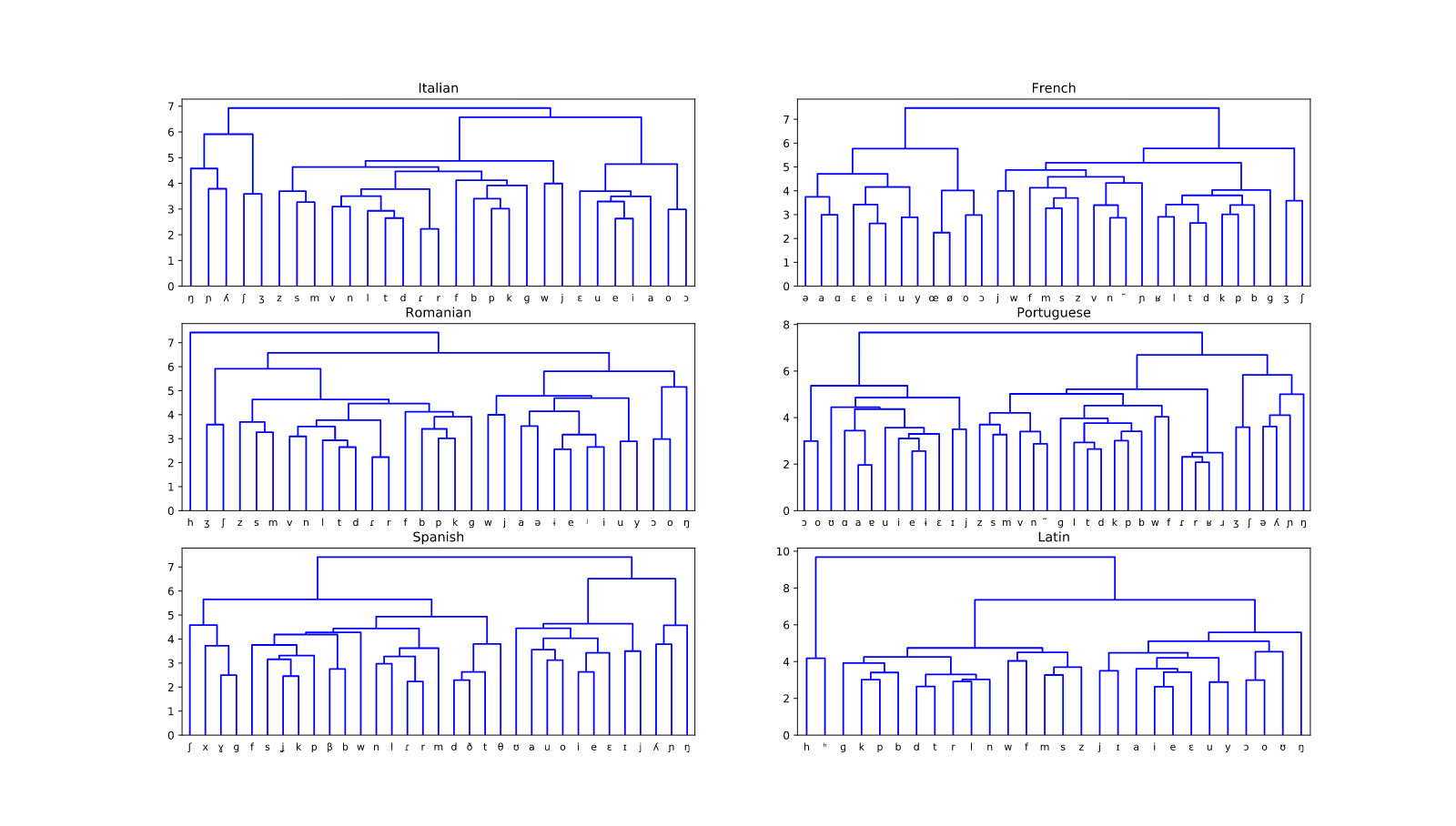}
    \vspace{-4em}
    \caption{}
     \label{appendix:clust-all}

\end{figure}

In this appendix, we show the hierarchical clustering created for all the languages in our dataset.
As it can be noted from figure \ref{appendix:clust-all}, the results for the different languages exhibit representations similar to those found in the French clustering: the primary division in each language is between vowels and consonants. In Portuguese, Latin, Spanish and Romanian some consonants are grouped together with vowels. These consonants are restricted to nasals, liquids or glides. The inclusion of these consonants can be explained by the peculiarity of their nature: all of them have a special phonological status, displaying similarities in their behavior to vowels.
In all languages phonologically related phonemes tend to be group under the same nodes. Among the others, glides are either found together with each other (as in French, Italian and Romanian) or with their vocalic counterparts (Latin, Spanish and Portuguese), consonants differentiated only in voicing are usually paired (\begin{IPA}[S]\end{IPA} and \begin{IPA}[Z]\end{IPA}), front and back vowels forms clusters and allophones usually shares the same node (Italian, French, Romanian, Spanish, Portuguese).

\end{document}